# oneDNN Graph Compiler: A Hybrid Approach for High-Performance Deep Learning Compilation

Jianhui Li
*Intel, US*
jian.hui.li@intel.com

Zhennan Qin
*Intel, China*
zhennan.qin@intel.com

Yijie Mei
*Intel, China*
yijie.mei@intel.com

Jingze Cui
*Intel, China*
jingze.cui@intel.com

Yunfei Song
*Intel, China*
yunfei.song@intel.com

Ciyong Chen
*Intel, China*
ciyong.chen@intel.com

Yifei Zhang
*Intel, China*
yifei.zhang@intel.com

Longsheng Du
*Intel, China*
longsheng.du@intel.com

Xianhang Cheng
*Intel, China*
xianhang.cheng@intel.com

Baihui Jin
*Intel, China*
baihui.jin@intel.com

Yan Zhang
*Intel, China*
yan3.zhang@intel.com

Jason Ye
*Intel, China*
jason.y.ye@intel.com

Eric Lin
*Intel, China*
eric.lin@intel.com

Dan Lavery
*Intel, US*
daniel.m.lavery@intel.com

***Abstract*—With the rapid development of deep learning models and hardware support for dense computing, the deep learning (DL) workload characteristics changed significantly from a few hot spots on compute-intensive operations to a broad range of operations scattered across the models. Accelerating a few compute-intensive operations using the expert-tuned implementation of primitives doesn't fully exploit the performance potential of AI hardware. Various efforts have been made to compile a full deep neural network (DNN) graph. One of the biggest challenges is to achieve high-performance tensor compilation by generating expert-level performance code for the dense compute-intensive operations and applying compilation optimization at the scope of DNN computation graph across multiple compute-intensive operations.**

**We present oneDNN Graph Compiler, a tensor compiler that employs a hybrid approach of using techniques from both compiler optimization and expert-tuned kernels for high-performance code generation of the deep neural network graph. oneDNN Graph Compiler addresses unique optimization challenges in the deep learning domain, such as low-precision computation, aggressive fusion of graph operations, optimization for static tensor shapes and memory layout, constant weight optimization, and memory buffer reuse. Experimental results demonstrate significant performance gains over existing tensor compiler and primitives library for performance-critical DNN computation graphs and end-to-end models on Intel® Xeon® Scalable Processors.**

## I. Introduction

With the rapid development of AI applications, deep learning software stacks and hardware are rapidly evolving. Data scientists are continuously exploring new deep neural network (DNN) models to improve the accuracy of the models by increasing the model parameters, using larger training datasets, and exploring innovative DNN structures and operations. Deep learning frameworks, like TensorFlow [3] and PyTorch [4], have been developed to support the development and deployment of deep learning models. While supporting Data scientists to develop new models, DL frameworks also need to efficiently use hardware resources to meet the huge computing needs of deep learning. Meanwhile, various types of hardware support for deep learning have been introduced, including adding matrix operation units on GPU and CPU, and hardware accelerators dedicated to deep learning computation.

Deep learning (DL) frameworks provide a rich set of deep neural network (DNN) operations for developers to describe a DNN model and use primitives libraries by default to offload the most performance-critical operations to CPU and GPU [1][2][20]. Most of the execution time of DL applications is spent on the DNN model. DL frameworks represent the DNN models internally as a computation graph of DNN operations. After performing high-level graph optimizations, the graph is traditionally executed operation by operation. On top of their own implementation of DNN ops, DL frameworks use third-party primitives libraries to offload the most performance-critical DNN operations.

Primitives library offers a simple and effective way to offload deep learning computation. However, with the fast evolution of AI software and hardware, the DL workload characteristics have been shifted from a few hot spots of concentrated compute-intensive operations to many scattered DNN operations, and the percentage of memory-bound operations is becoming significantly larger. Multiple sources contribute to the increasing time percentage of memory-bound operations. First, the deep neural networks used for natural language processing [19] and recommendation systems [18] have smaller input data and overall lower compute intensity compared to computer vision models [22][23]. Second, instead of supporting each innovative activation function with a complex DNN operation, DL Frameworks tend to compose multiple existing fine-grain operations, to maintain a balance of scalability and usability. Lastly, hardware acceleration usually focuses on accelerating the dense computation of low-precision data types and relies on the software to optimize memory-bound operations.

This performance characteristic shift drives the development of low-level graph compilers in the deep learning domain, also known as tensor compilers [5][6][7][10][15][17]. DL frameworks like TensorFlow and PyTorch capture and optimize users' input DNN graph and lower it to a low-level DNN graph with a reduced set of basic operations. The tensor compilers take the low-level DNN graph as input and focus on generating highly efficient code with optimization techniques specific to deep learning. Tensor compilers view DNN operations as tensor computation, internally represented as nested multi-level loops with the innermost loop body processing each tensor element. They use compiler loop transformation techniques to parallelize, vectorize, reorder, and merge the nested loops. XLA [5] fuses consecutive memory-intensive operations into one function and generates the code as one kernel. Triton [6] provides a tile-based programming model and focuses on compiling tile programs to efficient GPU code.

It has been a hot research area on how to automatically generate code for DNN computation graphs containing both compute-intensive and memory-intensive operations and achieve high-performance implementation. Tensor Comprehensions [9] provided a high-level language describing a DNN operation's mathematics and a polyhedral JIT compilation approach. Stripe [28] uses a Nested Polyhedral Model to enable automatic code generation for DNN operations. Various compiler techniques for loop parallelization and transformation trying to reach performance parity with the expert-tuned implementation have been explored using MLIR as an internal representation [6] [14] [21], with the goal of trying to reach performance parity with the expert-tuned implementation. Due to the inherent complexity of nested loop transformation, some tensor compiler researchers use autotuning methods to search for an optimal solution in a large search space [16] [25][27]. Performance library developers tend to use analytical models or cost model based heuristics to determine optimal tuning parameters for generating high-performance primitives [12]. The tensor compiler is mainly used in specific use cases where the expert-tuned primitives library can't offer the required performance and/or the users are willing to spend extra resources to find a better solution, very often via autotuning.

TVM [7] adopts Halide's approach to separate compute and schedule to represent DNN operations. The compute describes the nested multi-level loop, and the schedule represents possible transformations like loop tiling and reordering to find the best implementation on a target device. The optimization includes tensorization that maps instruction sequences or innermost loops to hardware-specific matrix instructions. TVM has developed an automated schedule optimizer that iteratively evaluates loop schedule proposals until an optimal one is found. Both TVM and oneDNN Graph Compiler take a DNN computation graph as input, perform both graph-level and operator-level optimization, and generate highly efficient low-level code for compute-intensive operations and fusion with neighboring memory-intensive operations.

oneDNN Graph Compiler is an open-source tensor compiler that automates the code generation for a DNN computation graph and is implemented as an embedded compiler component of oneDNN performance library. The goal of providing acceleration on top of existing deep learning frameworks and compilers has driven the major design decisions about optimizations and supporting IRs. In order to achieve the same level of computing efficiency for compute-intensive operations as primitives library implementation [1], oneDNN Graph Compiler develops target-specific templates and microkernels to generate expert-level kernels and fuse neighboring operations. Since the hardware type is limited, we believe it is a reasonable design choice that sacrifices some generality on kernel algorithm description in exchange for more direct control to achieve the best performance on a specific hardware device.

The contributions of the paper are the following:

- We propose a tensor compiler design with two level IR. Graph IR supports graph transformation like low-precision computation transformation, constant weigh preprocessing, layout propagation, and fine-grain fusion region formation. Tensor IR supports parallel-for loop generation for fine-grain fusion, merging of multiple Fused OP to one parallel loop for coarse-grain fusion, and memory buffer optimization.
- We introduce a template-based lowering for compute-intensive operations, which implements the best-known algorithm learned from expert-tuned kernel. The template uses microkernel and blocked layout and avoids lengthy and difficult compiler transformation passes to generate expert-level primitives.
- We introduce fused operation template to generate efficient code for fine-grain fusion. The template provides multiple anchor points for fusions so the compiler can choose the best places and assemble a parallel-for loop merged with loops representing neighboring fusible ops. It allows a broad set of flexible fusions including reduction and reordering operations without the complexity of parallel loop merging at low-level IR using traditional compiler techniques.

oneDNN Graph Compiler applies domain-specific expert knowledge that was distilled from the expert-tuned kernel development process to an automated compilation process and achieves comparable performance [1][11][12][13]. It combines compiler and kernel library techniques and focuses on domain-specific optimization problems. With expert-tuned microkernels and two levels of compiler IR, oneDNN Graph Compiler addresses domain-specific optimization challenges, such as generating efficient code with blocked data layouts specialized for static tensor shapes, constant weight optimization, aggressive fusion, and memory buffer reuse. On top of that, it further explores more advanced optimization at the graph level, such as optimizing the whole Multilayer Perception (MLP) network construct containing multiple matrix multiplication operations. Experimental results show that oneDNN Graph

Compiler delivers significant performance gains over primitive-based optimization for performance-critical DNN computation graph on CPU.

## II. High-level Design

A few important design choices allow us to replicate the performance of expert-tuned kernels and further attain superior graph-level performance with a manageable development effort. First, instead of lowering graph IR to a general nested loop representation and applying advanced loop transformations using compiler techniques, we use templates to mechanically generate code for compute intensive kernels and fusion with neighbor memory intensive operations. Second, instead of using multiple level IRs and gradual lowering, we choose two level IRs: Graph IR, and Tensor IR. Graph IR keeps OP semantics and supports graph optimizations, and Tensor IR abstracts hardware targets and supports low level optimization. Third, we use the microkernel to hide implementation details of the best-performant matrix instruction sequence. The templates and microkernel inherit the algorithm and implementation from the expert-tune kernel, and Graph and Tensor IR support graph-level optimization and potential reuse for other hardware targets.

Fig 1. provides a high-level view of oneDNN Graph Compiler design. The input DNN computation graph is internally represented as Graph IR. The Graph IR optimization module performs a number of transformations that optimize and group the computation graph as a sequence of fused operations. Graph IR is further lowered to Tensor IR. The Tensor IR doesn't preserve DNN operation semantics and is close to the C program semantics. The data structure it operates on is multidimensional arrays, representing tensor buffers in physical memory. Tensor IR is then further lowered to LLVM IR and intrinsic calls to Microkernels.

Using templates and microkernels greatly simplifies the compiler design. Just like implementing high-performance primitives using C language, the templates implement the best-known algorithm and can be instantiated with parameters and lowered to Tensor IR-based high-performance primitives. Since oneDNN Graph Compiler also uses templates to support fusing with neighboring operations, it doesn't require any sophisticated loop transformation and supporting loop IRs like MLIR Affine and Linalg dialects. Besides, by using microkernels, oneDNN Graph Compiler avoids developing low-level optimization techniques to generate efficient code at the instruction level.

Graph IR increases the optimization scope from individual primitives to a larger subgraph with multiple compute-intensive operations. As Graph IR retains the DNN OP semantics, most domain-specific optimizations are done at this level. The DNN OP semantics are implemented by the templates, which directly guide the decisions of parallel task decomposition, loop scheduling and tiling, tensor memory layout, and how to fuse with neighbor operations. Graph IR uses graph, logical tensor, and OP to describe a computation graph. A graph contains a set of OPs and logical tensors. Each OP represents an operation in a computation graph. A logical tensor represents the tensor's metadata, like the element's data type, shape, and memory layout. OP has kind, category, attributes, and logical tensors for inputs and outputs.

Graph IR optimization module first decomposes complex OPs into basic DNN OPs. The complex DNN OPs are OPs with complex semantics which could be composed of simple fundamental operations like addition and multiplication. They are introduced by DL frameworks to support high-level DNN OP semantics for ease of programming, such as batchnorm, quantize, gelu, and many activation operations. The basic DNN OPs are categorized to be either Tunable OP or Fusible OP. Tunable OPs describe DNN operations that use tunable parameters to instantiate a pre-defined template to generate the best-performing code. The examples include compute-intensive operations like matmul. Fusible OP refers to operations that can be fused to Tunable OPs, such as element-wise operations, broadcast, reduction, and data movement operations.

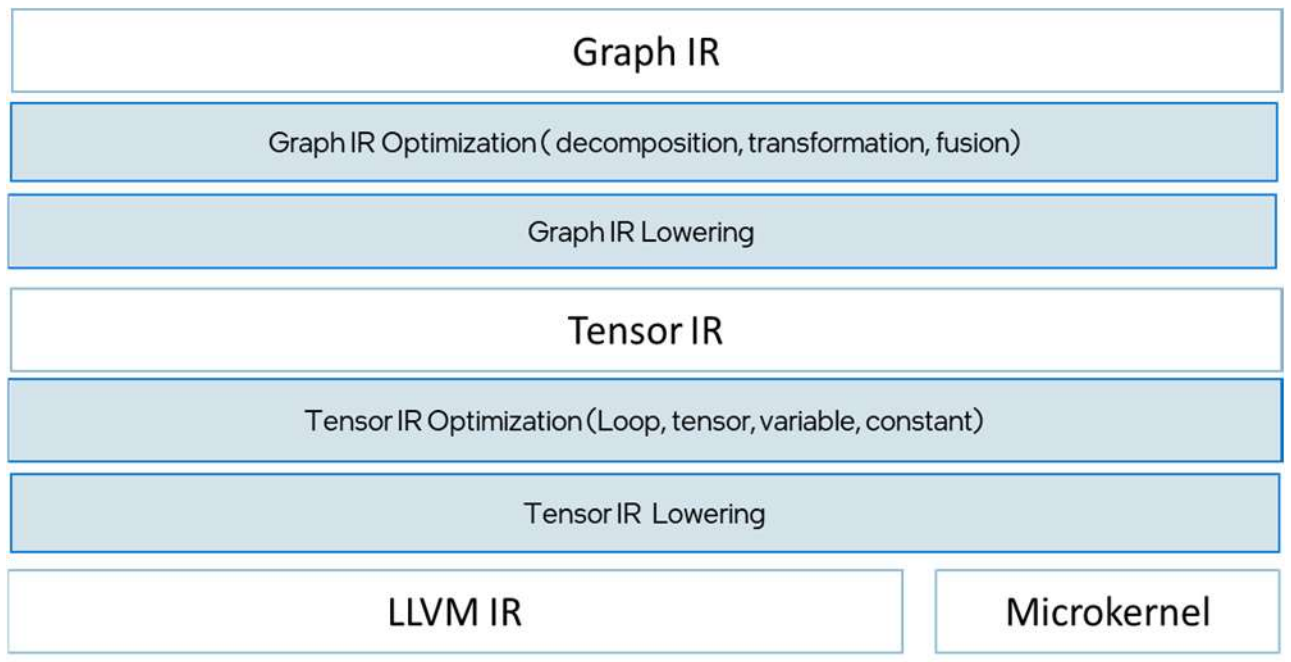

Fig. 1. oneDNN Graph Compiler IR and Optimization

The decomposition of complex DNN operations simplifies the Graph IR optimization module so it only needs to handle basic DNN operations. Besides the general compiler optimizations like common subexpression elimination (CSE), dead code elimination, and constant folding, it includes domain-specific optimizations like low-precision conversion, tensor memory layout propagation, constant weight preprocessing, and fusion. The fusion optimization pass decides whether it is profitable to fuse two operations and keeps fusing OPs to form a subgraph, which is represented as a Fused OP. The Graph IR is transformed into a graph of Fused OPs and then lowered into Tensor IR.

Tensor IR supports mechanically lowering Fused OP to simple loops, without complex nested loop analysis and transformation. Tensor IR optimization mainly focuses on tensor buffer optimization and supports low-level code generation. Just like the C program, Tensor IR supports function, statement, expression, and intrinsic functions. The Tensor IR module, lowered from a Graph IR graph, contains multiple functions, each of which represents a lowered Fused OP. The Tensor IR module has an entry function that contains a sequence of calls to other functions lowered from Fused OPs. A Tensor IR function contains multiple statements build on expressions, which operate on constants, variables, and tensors. Constants and variables represent individual data elements, used to represent scalar data like loop index, tensor shape, address, and offset to tensor buffer. Tensors represent multi-dimension arrays backed by a data buffer. The intrinsic function is used to represent a microkernel, which is carefully hand-tuned and fulfills a subtask of a DNN OP with data in the fastest cache on a single CPU core.

## III. MICROKERNEL-BASED TEMPLATE FOR TUNABLE OP LOWERING

Automating the high-performance code generation for Tunable OPs is the foundation of a tensor compiler. oneDNN Graph Compiler took an approach inherited from the performance library development, which first creates the code templates for a given Tunable OP and then instantiates it with parameters decided by a heuristic. The parameters are decided based on the input data tensor shape and hardware sizes of the microarchitecture.

The template shown in Fig 2. is for a matmul op that does matrix multiplication over A[M, K] and B[K, N] and produces C[M, N]. The template is applied to a common deep learning use case where the computation uses multiple cores, and the size of input and output tensor fits within the cache system. The outer parallel loops divide the kernel into multiple subtasks for multi-cores. Each subtask is assigned to one single core, named single-core kernel, which is represented by the inner loops which call a microkernel in the innermost loop body.

The microkernel and the single-core kernel operate on a tensor slice that represents a subset of tensor elements. For example, the original tensor is represented as A[0:M, 0:N], where the subscription represents starting offset and size for each dimension. The tensor slice is represented as A[0:MB, 0:NB], where MB and NB refer to the tile size of the tensor slice along m and n dimensions. A submatrix is a special case of a 2-dimension tensor slice. In the template above, the microkernel produces a small submatrix C[0:MB, 0:NB], and the single-core kernel outputs a larger submatrix C[0:MSN, 0:NSN].

The microkernel is an important element for the oneDNN Graph Compiler to achieve comparable performance to expert-tuned primitives. oneDNN Graph Compiler uses the microkernel named batch-reduce GEMM [8][24]. The microkernel has two inputs, both representing a batch of 2D matrices. It first applies matrix multiplication with each batch element to produce a batch of immediate 2D matrices and then sums them to a final 2D matrix output. This interface can be used for many variants of matmul and convolution op in both inference and training use cases and was adopted by both oneDNN primitives and oneDNN Graph Compiler.

The microkernel is fine-tuned to maximize the compute efficiency by fully utilizing the compute function unit and the high bandwidth provided by registers and the L1 cache. It abstracts the ISA difference so oneDNN Graph Compiler doesn't need to deal with different vector or matrix instructions provided by different CPUs. However, the oneDNN Graph Compiler needs to choose the input submatrix sizes for the microkernel so that they are usually multiples of register sizes used by the vector and matrix function units. Also, it needs to choose the batch size for the microkernel so that the whole input and output submatrices fit within the L1 cache. To further streamline the cache access, the input and output tensors are blocked. To simplify the implementation, the input and output tensors are blocked using the submatrix sizes [MB, NB, KB]. So, each microkernel accesses a contiguous memory buffer.

The parameters for lowering a matmul op refer to the variable values in the template above: MPN, NPN, MB, NB, KB, BS, and ordering of loops indexed by msi, ksi, and nsi. The other parameters can be derived from the parameters above. oneDNN Graph Compiler uses an expert-tuned heuristic to decide these parameters. For a given output matrix size, it first proposes single-core kernel size options, a set of [MPN, NPN], which can use all cores with good load balance. It further proposes microkernel size options, a set of [MB, NB, KB, BS], which ensure good microkernel performance. Then the heuristic picks a pair of these sizes, which has the best overall kernel performance for the entire system with all cores. It iteratively searches for the best parameters, based on a cost model which considers multi-core load balancing and single-core kernel efficiency. Heuristic also reports the loop ordering of the inner loops which it assumes when computing the single-core kernel efficiency during the search process.

oneDNN Graph Compiler developed multiple templates for different uses. One Tunable OP can have multiple templates depending on the use cases. For example, for inference cases, sometimes the use case only processes one data sample with multiple cores so that the template may have to apply "k-slicing" to extract additional parallelism from the reduction axis.

multi-core kernel | single-core kernel | micro kernel

```
Parallel loop mpi = 0, MPN, 1 {
  Parallel loop npi = 0, NPN, 1 {
    Loop msi = 0, MSN, 1 {
      mpsi = mpi * M/MPN + msi;
      C'[0:NSN, 0:MB, 0:NB] = 0;
      Loop ksi  = 0, KSN, BS {
        Loop nsi = 0, NSN, 1 {
          npsi = npi*N/NPN + nsi;
           A_addr[0..BS] = &A[mpsi:1, ksi:0..BS, 0:MB, 0:KB];
           B_addr[0..BS] = &B[ksi:0..BS, npsi:1, 0:NB, 0:KB];
          C'[npsi:1,0:MB, 0:NB] += Batch_reduce_gemm
                   (A_addr[0..BS], B_addr[0..BS],Batch = BS);
        }
      }
      C[mpsi:1, npi:NSN, 0:MB, 0:NB] = C'[0:NSN, 0:MB, 0:NB];
    }
  }
}
```

| | m | n | k |
|---|---|---|---|
| Index of single-core kernel within multi-core kernel | mpi | npi | kpi |
| Number of single-core kernels within multi-core kernel | MPN | NPN | KPN |
| Index of microkernel within single-core kernel | msi | nsi | ksi |
| Number of microkernel within single-core kernel | MSN | NSN | KSN/BS |
| Index of microkernel within multi-core kernel | mpsi | npsi | kpsi |
| Number of microkernel within multi-core kernel | MPSN | NPSN | KPSN/BS |
| Tensor size | M | N | K |
| Tensor block size | MB | NB | KB |
| Tensor slice size accessed by Microkernel size (batch size = BS) | MB | NB | KB * BS |
| Tensor slice size accessed by single-core kernel | MSBN = MB * MSN | NSBN = NB * NSN | KSBN = KB * KSN |
| Tensor slice size accessed by multi-core kernel | M = MB * MSN * MPN = MB * MPSN | N = NB * NSN * NPN = NB * NPSN | K = KB *KSN * KPN = KB * KPSN |

Tensor is described with a Tensor name followed by index and size for each dimension. Tensor A[0:M, 0:K] refers to 2 dimensions tensor starting from the position [0,0] with size [M, K]. A[0:MB, 0:KB] refers to a tensor slice containing a subset of A tensor elements, starting from position 0 to MB-1 along the m dimension, and 0 to NB-1 along the n dimension. The pseudo-code uses a blocked layout for A, B, and C. C[0:MPSN, 0:NPSN, 0:MB, 0:NB] denotes the full C tensor C[0:M, 0:N] reordered with a blocked layout. C[mps:1, np:NSN, 0:MB, 0:NB] denotes a tensor slice which "slice" the C tensor in the first 2 dimensions starting from position "mps" and "np" with size "1" and "NSN". A_addr[0..BS] denotes an array with BS elements from A_addr[0] to A_addr[BS-1]. A[mps:1, ks:0..BS, 0:MB, 0:KB] denotes an array of BS tensor slices from A[mps:1, ks:0, 0:MB, 0:KB] to A[mps:1, ks:BS-1, 0:MB, 0:KB].

Fig. 2. Microkernel based template for Tunable OP

## IV. Template with Anchors for Fused OP Lowering

oneDNN Graph Compiler combines a Tunable op with multiple adjacent Fusible ops to a Fused op and lowers it to a nested loop using the Fused OP template. The template contains placeholders, known as anchors, at the beginning and the end of each loop level for the input and output tensors. The Graph IR fusion optimization decides whether it is profitable to fuse a Fusible op to a Tunable op and which anchor point is assigned to the Fusible op. The Fused OP lowering pass retrieves anchors for Fusible ops and directly inserts its corresponding Tensor IR at the anchor.

Fig. 3 illustrates the anchors within a template and the associated tensor slices for each anchor. The anchors preceding the microkernel are referred to as pre-op anchors, while those following the microkernel are termed post-op anchors. The right table in Fig. 3 shows the tensor slice working set size for each anchor point which describes the memory size accessed by the fused operation at the anchor point on a single core. It also shows the formula to compute how many times the fused op is invoked within a single-core kernel and how many total tensor element memory accesses are needed for each anchor point. The concrete number can be deduced when the template is instantiated with the parameters for a Tunable op.

The fusion optimization uses a heuristic to decide which anchor to choose. The heuristic evaluates the cost of a single-core kernel between all possible anchors and the option of not fusing, and then it chooses the one with the lowest estimated cost. The commit anchors inside the innermost loop work on the smallest tensor slice, which provides a low per-access cost as the data is in the fastest cache. So the post-op usually finds the first anchor point toward the innermost loop the best choice. However, for pre-op, the Fused OP lowering considers both the computation and temporary buffer size introduced by pre-op fusion. The anchors at inner loop bodies require smaller temporary buffer size but may have redundant computations which can be avoided by careful selection of anchor points.

```
Parallel loop mp i= 0, MPN, 1 {
  Parallel loop npi = 0, NPN, 1 {
    Loop msi = 0, MSN, 1 {
      mpsi = mpi * M/MPN + ms;
      C'[0:NSN, 0:MB, 0:NB] = 0;
       Loop ksi = 0, KSN, BS {
         Reorder(A, [1, 1], A'[mpsi:1, ksi:BS, 0:MB, 0:KB],
                                   [MB, KB], from=[mpsi, ksi]);
         Loop nsi = 0, NSN, 1 {
           npsi = npi * N/NPN + nsi;
           A'_addr[0:BS] = &A'[mpsi, ksi:BS, 0, 0];
           B_addr[0:BS] = &B[ksi:BS, npsi, 0, 0];
           C'[nsi:1,0:MB, 0:NB] += Batch_reduce_gemm
                   (A'_addr[0:BS], B_addr[0:BS],Batch = BS);
         }
      }
      C''[mpsi:1, npsi:NSN, 0:MB, 0:NB] = C'[0:NSN, 0:MB, 0:NB];
      C'''[mpsi:1, npi:NSN, 0:MB, 0:NB]) =
                             Relu(C''[mpsi:1, npi:NSN, 0:MB, 0:NB]);
      Reorder( C'''[mpsi:1, npi:NSN, 0:MB, 0:NB], [MB, NB],
                                      C, [MB2, NB2], to=[mpsi, npi]);
    }
  }
}
```

Fig. 4. Pseudo code for Fused OP

Fig. 4 shows a pseudo-code for fusing data layout reorder and ReLU (rectified linear unit) ops to an instantiated GEMM op. The first reorder op is inserted as pre-op fusion at anchor #4, which converts from a plain layout tensor A to a blocked layout A' with blocking factors MB and KB. The fused reorder op works on the tensor slice of A', denoted as A'[mpsi:1, ksi:BS, 0:MB, 0:KB], which starts from the position A'[mpsi, ksi, 0, 0] and has a slice with the size of [BS, MB, KB]. It also fuses two post-ops, a ReLU op followed by a reorder op. Both operations are inserted at the post-op anchor #1. The reorder op changes the memory layout of the C tensor from the blocking factor of MB and NB to MB2 and NB2.

```
Parallel loop mpi = 0, MPN, 1 {
  pre_op_anchor#1 : A[mpi*MSN:MSN, 0:KSN, 0:MB, 0:KB];
  pre_op_anchor#1 : B[0:KSN, 0:NPSN, 0:NB, 0:KB];
  Parallel loop npi = 0, NPN, 1 {
    pre_op_anchor#2 : A[mpi*MSN:MSN, 0:KSN, 0:MB, 0:KB];
    pre_op_anchor#2 : B[0:KSN,npi*NSN:NSN, 0:NB, 0:KB];
    Loop msi = 0, MSN, 1 {
      mpsi = mpi * M/MPN + msi;
      pre_op_anchor#3 : A[mpsi:1, 0:KSN, 0:MB, 0:KB];
      pre_op_anchor#3 : B[0:KSN,npi*NSN:NSN, 0:NB, 0:KB];
       C'[0:NSN, 0:MB, 0:NB] = 0;
       Loop ksi = 0, KSN, BS {
         pre_op_anchor#4 : A[mpsi:1, ksi:BS, 0:MB, 0:KB];
         pre_op_anchor#4 : B[ksi:BS, npi*NSN:NSN, 0:NB, 0:KB];
         Loop nsi = 0, NSN, 1 {
           npsi = npi*N/NPN + nsi;
           pre_op_anchor#5 : A[mpsi:1, ksi:BS, 0:MB, 0:KB];
           pre_op_anchor#5 : B[ksi:BS, npsi:1, 0:NB, 0:KB];
           A_addr[0..BS] = &A[mpsi:1, ksi: 0..BS, 0:MB, 0:KB];
           B_addr[0..BS] = &B[ksi:0..BS, npsi:1, 0:NB, 0:KB];
           C'[nsi:1,0:MB, 0:NB] += Batch_reduce_gemm
                 (A_addr[0..BS], B_addr[0..BS],Batch = BS);
         }
      }
      C[mpsi:1, npi:NSN, 0:MB, 0:NB] = C'[0:NSN, 0:MB, 0:NB];
     post_op_anchor#1 : C[mpsi:1, npi:NSN, 0:MB, 0:NB];
   }
   post_op_anchor#2 : C[mpi*MSN:MSN, npi*NSN:NSN, 0:MB, 0:NB];
 }
 post_op_anchor#3 : C[mpi*MSN:MSN, 0:NPSN, 0:MB, 0:NB];
}
```

| Anchor | Tensor slice's working set size per core | Access times per core | Total memory access per core |
|---|---|---|---|
| pre_op_anchor#1 | A' [MSN, KSN, MB, KB]<br>B' [KSN,NPSN, NB, KB] | 1 | MSN* MB * KSN * KB<br>NPSN * NB * KSN * KB |
| pre_op_anchor#2 | A' [MSN, KSN, MB, KB]<br>B' [KSN,NSN, NB, KB] | 1 | MSN* MB * KSN * KB<br>NSN * NB * KSN * KB |
| pre_op_anchor#3 | A' [KSN, MB, KB]<br>B' [KSN, NSN, NB, KB] | MSN | MSN * MB * KSN * KB<br>MSN * NSN * NB * KSN * KB |
| pre_op_anchor#4 | A' [BS, MB, KB]<br>B' [BS, NSN, NB, KB] | MSN * KSN/BS | MSN * MB * KSN * KB<br>MSN * NSN * NB * KSN * KB |
| pre_op_anchor#5 | A' [BS, MB, KB]<br>B' [BS, KB, NB] | MSN * NSN * KSN/ BS | MSN * MB * KSN * KB *NSN<br>MSN * NSN * NB * KSN * KB |
| post_op_anchor#1 | C[MB, NSBN] | MSN | MSBN*NSBN |
| post_op_anchor#2 | C[MSBN, NSBN] | 1 | MSBN*NSBN |
| post_op_anchor#3 | C[MSBN, N] | 1 | MSBN * N |

The template has predefined anchors as placeholders to fuse pre-ops and post-ops. Each anchor point is associated with a tensor slice for the pre-ops and post-ops to work on. Once the blocking parameters are decided, the tensor slice size and access times can be deduced to support the fusion decision. The fusion optimization pass chooses anchor points for groups of pre-ops and post-ops according to the estimated computation cost.

Fig. 3. Fused OP template with anchors and cost table

## V. GRAPH IR OPTIMIZATION

oneDNN Graph Compiler Graph IR optimization transforms the graph to use low-precision computation and preprocessed constant tensors, and then prepares the graph as a sequence of Fused ops for optimized code generation. The Graph IR is first decomposed into a graph of basic DNN operations to simplify the optimization passes, and clustered to form fine-grain Fused ops, and then lowered to Tensor IR using the Fused OP templates.

Fig. 5 illustrates these optimization passes with a quantized multilayer perceptron (MLP) example. The input DNN quantized MLP in Fig5.1 contains two matmul ops, and the activation ops are omitted for simple illustration. Each FP32 matmul op is surrounded by two dequantize ops and a quantize op, denoted by DQ and Q. The dequantize converts an Int8 data type tensor to FP32 and the quantize op does the reverse. It uses the asymmetric quantization scheme, so the first dequantize op scales A input tensor by a_s and then offset by a_z to adjust the zero point, and the other dequantize op scales the weight matrix B with b_s. The optimized DNN Graph in Fig5.2 shows the effect of low-precision conversion optimization. It first breaks down the quantize and dequantize op to be simple addition and multiply ops and transform the graph to be a mathematically equivalent form that uses Int8 matmul op. The low-precision conversion brings significant speedup as it reduces both the computation and memory bandwidth required to compute a deep learning model.

Fig 5.2 also illustrates the effects of constant weight preprocessing optimization, which recognizes the constant weight tensor and its related computation and builds a special initial function that preprocesses the constant weight and reuses the preprocessed weight at the runtime. For the static quantization inference use case, the weight tensors and quantization parameters are constant, so the computation over constant weight, scale, and zero point can be avoided completely at runtime. As the weight data buffer might not be available during the compilation, so the compiled code needs to generate an function to preprocess the constant weight at the execution time when it first arrives. The quantization parameters, a_s, b_s, c_s, and c_z are constants passed as dequantize op's attribute, which can be folded to the generate code when lowering to Tensor IR.

The fine-grain fusion optimization clusters the graph to fine-grain graph regions and encapsulates them as Fused ops. It first considers the immediate succeeding ops of the Tunable op as post-op candidates and keeps growing the Fused OP region. The post-op could be elementwise, broadcast, reduction, and reorder ops, and multiple post-ops may be added to a Fused OP region. For example, the activation and normalization ops after matmul op are broken down to basic ops and added to the Fused OP region. The region stops growing when a limit is reached, say, the post-op sequence can only have one reorder, one reduction, certain number of total ops, or total size of additional inputs. Then it looks for preceding ops as pre-op candidates. The pre-op fusion only supports limited cases like reorder and transpose operations and only be used at the entry point of the graph. For a Fusible op between two Tunable ops, it is typically more profitable to fuse as post-op of the first Tunable op, so the fusion optimization first adds post-op and then pre-op to Fused OP region.

The layout propagation optimization exploits extra performance benefits cross Tunable ops by allowing Tunable op to use the most desired blocked layout. As Tunable op relies on the blocked layout to achieve the best performance on the CPU, very often the best-performed blocked layout might be different between two Tunable ops. It allows the Tunable ops within a graph to use a blocked layout but keep the graph input/output tensor as a plain layout. It first inserts reorder operations at the graph boundary to ensure the entry and exit points using the plain layout. Then it iterates the DNN computation graph and inserts reorder operation between two Tunable ops if they use different blocked layouts. It first queries a Tunable op for its desired blocked layouts, if none of the desired blocked layouts is consistent with the current layout, it inserts a reorder op before the Tunable op. Fig. 5.3 illustrates the fine-grain fusion region and newly inserted reordered ops at the boundary of graph and internally between Tunable ops. The reorder op between two Tunable ops is added to the end of the Fusion OP region of the previous Tunable op. The reorder for the input constant weights is converted to a preprocess weight, named prepacked weight, using the constant weight preprocessing optimization.

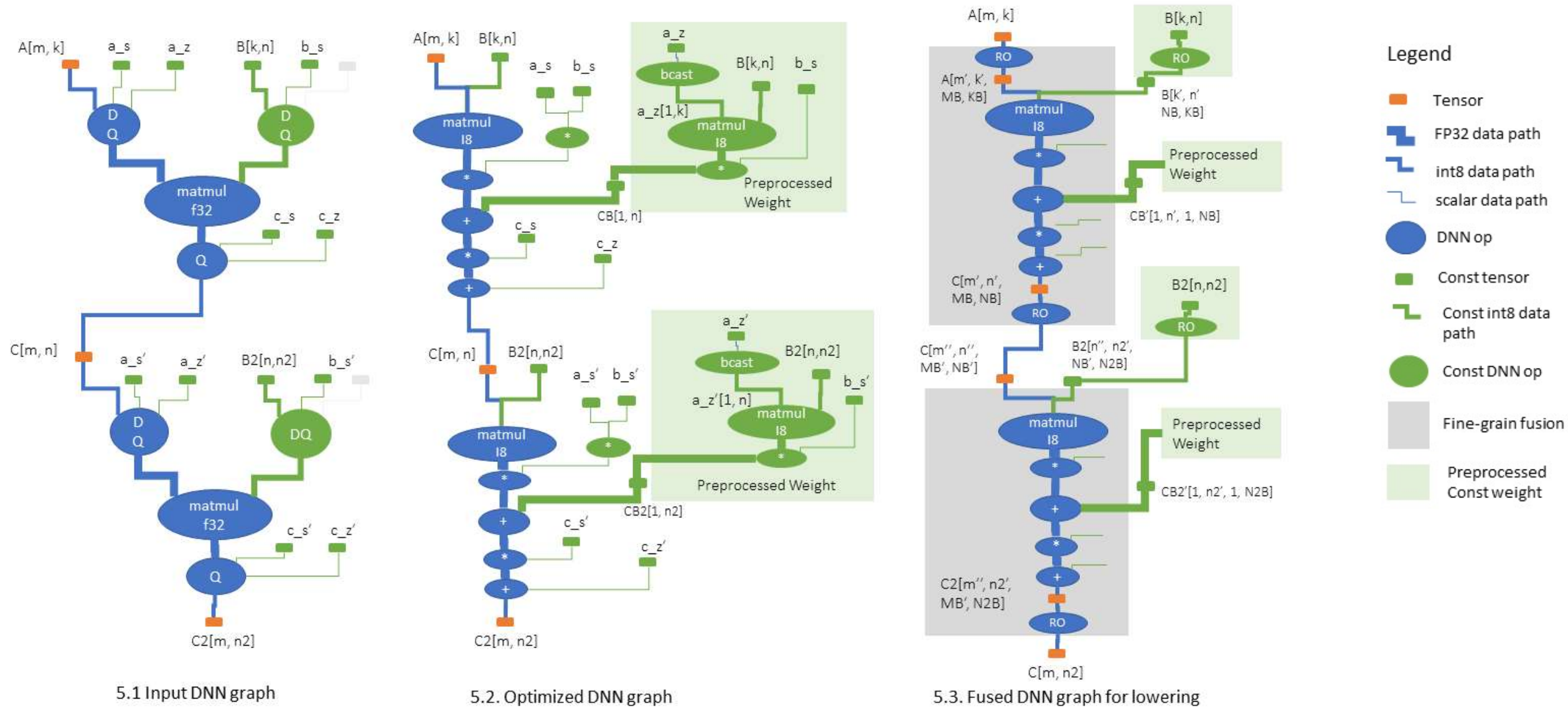

Fig. 5. Graph IR optimization

The coarse-grain optimization happens at the final stage of optimizations, where the graph is converted to a list of Fused ops in topological order and lowered to a sequence of parallel-for. It merges multiple parallel-for loops lowered form Fused ops together to one parallel-for loop. The mechanism of merging of multiple parallel-for is supported at the Tensor IR level, but the merging decision is done at the Graph IR level as part of lowering. The coarse-grain optimization greatly improves data locality across two Fused ops and can be applied to many graph patterns. For example, as the two matmul ops in MLP use the same batch, the lowered two nested parallel loops have the same outermost loop iterating the batch dimension, which can be further merged as one parallel loop. When the heuristic chooses the parameters for a Fused op, it tries to choose the outermost loop blocking factor best aligned with core numbers, so the instantiated fused op likely has the same blocking factors as its neighbor. When the coarse-grain fusion optimization decides to merge two fused ops, it marks the two nested loops in Tensor IR as "mergeable" during the lowering process. Then Tensor IR merges two nested loops mechanically as guided by the Graph IR optimizations.

## VI. TENSOR IR OPTIMIZATION

Tensor IR is the lowest intermediate representation in the oneDNN Graph Compiler. At the Tensor IR level, the DNN computation graph is lowered to a C-like program, which includes function, statement, expression, and intrinsic functions. The Fused op is lowered as a function, which contains nested loops. A complex statement describes a structure like a loop, and a simple statement does computation. Var and Tensor represent scalar variables and multi-dimension arrays respectively.

Tensor IR supports Graph IR optimization by merging loops as instructed by Graph IR. Fig. 6 shows the example of Tensor IR for the pseudo-code in Fig. 4. In the Tensor IR, the computation on the tensor slices is represented by either a nested loop or a function call to the microkernel. The inserted pre-op and post-op are lowered to nested loops. The two post-op, ReLU and reorder ops, are merged as one nested loop using the hint passed by Graph IR.

The main optimizations on Tensor IR are tensor size optimization and memory buffer optimization. Tensor size optimization tries to reduce the tensor size of each temporary tensor. The temporary tensors are introduced in the pre-op and post-op fusion process. The temporary tensor was initially introduced as a full-size tensor in the lowering process and then reduced by the tensor size optimization. In the example code of Fig. 6, the post-ops are fused into one loop nest in the Tensor IR. Since the accesses of the temporary tensors, C'' and C''', are local to the innermost loop body, the temporary tensor could be replaced by a scalar variable for a smaller memory footprint and better cache locality. The temporary tensor introduced by pre-op fusion can be reduced similarly by analyzing the scope of the tensor usage. For example, A'[MSN, BS, MB, KB] could be reduced to A' [BS, NB, KB], since the producer of A' and consume are within the "msi" loop, so there is no need to save the result along the 2nd dimension of A'.

After tensor size optimization, the multiple-dimension tensor representation is flattened to a one-dimensional array to represent the memory buffer. The memory buffer optimization tries to reuse the memory buffer of temporary tensors to have a minimum overall temporary buffer size for the compiled code and tries to improve the locality of the temporary buffer use. The main target of memory buffer optimization is to reuse the memory buffer created for the temporary tensors between fused op. In the inference use case, the output tensor is only consumed by the next fused op, and so the buffer could be reclaimed once the next fused op completed execution. Since the input tensor size is known to the compilation process, the internal memory buffer usage can be calculated at the compile time and optimized to improve efficiency.

Memory buffer optimization uses life span analysis like traditional compiler analysis for register allocation based on the def-use chain. The algorithm considers both reusing the hot memory and reducing the overall peak memory. At each point, when an intermediate buffer is needed, it tries to reuse the free intermediate buffers, which are already allocated but not used anymore. Among multiple choices of reusable memory buffers, it chooses the one that was used most recently, so likely the data is still in the cache system.

```
Var Const MPN, NPN, MSN, NSN, BS, KSN, MB, NB;
Tensor FP32[M, K] A;
Tensor FP32[M/MB, K/KB, MB, KB] A';
Tensor FP32[K/KB, N/NB, NB, KB] B;
Tensor FP32[NSN, MB, NB] C';
Tensor FP32[M/MB,N/NB, MB, NB] C'', C''';
Tensor FP32[M/MB2, N/NB2, MB2, NB2] C;
Var int* A_addr[BS], B_addr[BS];
Var Index mp, np, ms, ks, ns, nps, mps;
Parallel loop mp i= 0, MPN, 1 {
  Parallel loop npi = 0, NPN, 1 {
    Loop msi = 0, MSN, 1 {
      mpsi = mpi * M/MPN + msi;
      Loop nsi = 0, NSN, 1 {
        Loop mbi = 0, MB, 1 {
          Loop nbi = 0, NB, 1 {
            C'[0:NSN, 0:MB, 0:NB] = 0;
      } } }
      Loop ksi = 0, KSN, BS {
          Loop bsi = 0, BS, 1 {
              ksbi = ksi * BS + bsi;
              Loop mbi = 0, NB, 1 {
                  Loop kbi = 0, KB, 1 {
                      A'[mpsi, ksbi, mbi, kbi]
                                 = A[mpsi*MB + mbi, ksbi*KB+kbi];
          }   }   }
          Loop nsi = 0, NSN, 1 {
              npsi = npi * N/NPN + nsi;
              Loop bsi = 0, BS, 1 {
                A'_addr[bsi] = &A'[mpsi, ksi, 0, 0];
                B_addr[bsi] = &B [ksi, npsi, 0, 0];
              }
              C'_addr = &C'[nsi,0, 0] ;
              Batch_reduce_gemm(A'_addr, B_addr,
                         C'_addr, MB, NB, KB, Batch = BS);
          }
      } //end of loop ksi
      Loop nsi = 0, NSN, 1 {
         npsi = npi * N/NPN + nsi;
         Loop mbi = 0, MB, 1 {
            Loop nbi = 0, NB, 1 {
               C''[mpsi, npsi, mbi, nbi] = C'[nsi, mbi, nbi];
               C'''[mpsi, npsi, mbi, nbi]= max(C''[mpsi, npsi, mbi, nbi], 0);
               C[(mpsi*MB+mbi)/MB2, (npsi*NB+nbi)/NB2,
                  (mpsi*MB+mbi)%MB2, (npsi* NB+nbi)%NB2]
                     = C'''[mpsi, npsi, mbi, nbi];
      } }  } // end of loop nsi
    } // end of loop msi
}  }  // end of parallel loop mpi, npi
```

Fig. 6. Example of Tensor IR

## VII. Experimental Results

oneDNN Graph Compiler is built as an embedded component of oneDNN library to accelerate the DNN computation subgraphs. oneDNN is the industry-standard best-performing library implementation and has been integrated into multiple DL frameworks as the default performance library to accelerate deep learning on the CPU. Besides primitives API, oneDNN provides a Graph API, so that DL frameworks or graph compilers can use it to accelerate the DNN computation subgraphs. Achieving the best execution efficiency for target DL workloads requires improvements from both DL frameworks and oneDNN Graph Compiler. DL framework developers first profile the execution and identify performance critical DNN operations and subgraphs. For the cases where oneDNN Graph Compiler provides performance benefits, DL framework passes the subgraphs to oneDNN Graph for acceleration. oneDNN Graph Compiler continuously enhances the templates, microkernels, heuristics, and fusion capability for the new subgraphs including adding operations and tuning for a broader range of data shapes. The current oneDNN Graph Compiler already provides well-implemented templates for widely used operators like matmul and convolution which can provide good performance for a wide range of shapes.

We choose BERT Large and DLRM, representing DNN models for natural language processing and recommendation systems, to demonstrate the performance benefit of oneDNN Graph Compiler for the inference use case with both FP32 and Int8 data types. There are two performance-critical DNN computation subgraphs in these two DNN models. The Multilayer Perceptron (MLP) contains multiple matmul ops intermixed with activation ops like ReLU. The MLP subgraph is the basic building block for many deep learning models. The Multi-Head Attention (MHA) subgraph is the key element to Transformer based deep learning models like Bert for natural language processing. The MHA subgraph referred in this paper focuses on the scaled dot-product attention portion of the whole MHA graph, which contains two batch matmul ops and a softmax as well as other binary ops between them.

Table 1 shows the problem sizes and data type we used for performance evaluation. We select a wide range of batch sizes and several representative data shapes for weights and input tensors. The weight sizes for MLP are from the MLPerf DLRM model, and the sequence length and hidden size choices for MHA are from Bert models. The Int8 quantization scheme uses u8 asymmetric quantization for activation and per channel s8 symmetric quantization for weight. The input and output matrices are in plain layouts. The performance data is collected on an Intel® Xeon® Platinum 8358 processor with 32 cores.

We construct MLP and MHA tests according to Table 1 and compare the performance of DNN computation subgraph with TVM and oneDNN primitives using the best-known method. As the input workloads are represented as graphs, we make sure to apply all necessary graph transformations. For TVM, we constructed a graph using Relay graph op and used an auto-scheduler and performed autotune for the best result. TVM is able to fuse memory-intensive operations to the matmul operation. We also compare with oneDNN expert-tuned primitives and post-op fusion implementation. oneDNN post-op API supports fusing matmul op with ReLU in the MLP tests, and matmul with division and addition ops in the MHA tests. The tests use low-precision post-op fusion API with prepacked and preprocessed compensated weight following best-known practice. For oneDNN Graph Compiler, we use oneDNN Graph API to construct the MLP and MHA test cases. For oneDNN Graph Compiler, weight prepacking and compensated weight preprocessing are done automatically without additional steps, and there is no need to autotune.

We first compare the individual matmul operation performance in the MLP tests between three implementations. Instead of running each individual matmul operation separately, our methodology is to run the matmul operations consecutively as an MLP subgraph in the end-to-end DL model, since the measurement of individual matmul operations does not correlate to the performance in the real workloads. Running consecutive matmul operations in the subgraph better emulates the cache locality effects. We also use TVM auto-scheduler to tune the matmul in the context of MLP to ensure the best TVM result. We run oneDNN Graph Compiler with the coarse-grain fusion disabled so that we can measure each operation separately.

The performance comparison result is presented as execution speedup over TVM baseline in Fig. 7. For FP32 matmul, we observed that TVM outperforms oneDNN Graph Compiler in 15 test cases, but its total execution time falls behind oneDNN Graph Compiler by 1.4x. The 15 test cases have very small computation sizes and account for less than 5% of the time of total test cases, so these test cases are less performance critical in the end-to-end DNN model execution. We identify two reasons that the oneDNN Graph compiler falls behind TVM. First, the last 5 test cases are GEMMV operations, which involve significant overhead for oneDNN Graph Compiler to pad the input matrix to match the minimum size required by the microkernel. Second, the kernel built by oneDNN Graph Compiler is always configured to use all available cores, because allowing matmul kernel to use fewer threads brought a negative performance impact when running end-to-end DNN models. However, this triggers significant synchronization overhead when each individual kernel is measured, especially when the computation size is small. By tuning the number of threads, we can reduce the gap to within 10% range of TVM performance.

TABLE I. Workload Parameters

| test name | data type | input batch size | sequence length | hidden size | head numbers |
|---|---|---|---|---|---|
| MLP-1 | Int8, FP32 | 32, 64, 128, 256, 512 | N/A | 13x512x256x128 | N/A |
| MLP-2 | Int8, FP32 | 32, 64, 128, 256, 512 | N/A | 479x1024x1024x512x256x1 | N/A |
| MHA-1 | Int8, FP32 | 32, 64, 128 | 128 | 768 | 8 |
| MHA-2 | Int8, FP32 | 32, 64, 128 | 128 | 768 | 12 |
| MHA-3 | Int8, FP32 | 32, 64, 128 | 384 | 1024 | 8 |
| MHA-4 | Int8, FP32 | 32, 64, 128 | 512 | 1024 | 16 |

For Int8 matmul, we observed that oneDNN Graph Compiler significantly outperforms TVM by 3.8x for the total test execution time. Compared to oneDNN Graph Compiler's 3.0x speedup of Int8 matmul over FP32, TVM's Int8 matmul execution time only improves by 11% over FP32. It appears that TVM's auto-scheduler can't find a good schedule to realize the full performance benefit of Int8 computation. First, the Int8 matrix multiplication is mapped to Int8 VNNI instruction which requires a special data layout reordering. In addition, it requires sophisticated blocking since the Int8 data type reduces the matrix data size by 4x, and fitting the data within L1 cache makes a significant performance difference. We believe the performance difference is mainly related to TVM implementation; however, it is not straightforward to develop and tune the baseline schedule so that auto-scheduler can search for the best schedule. Compared to TVM's schedule approach, oneDNN Graph compiler's template is closer to the generated kernel code, so the developers have better control over what the kernel looks like.

oneDNN primitives significantly outperform TVM by 1.4x for FP32 matmul test cases and 3.6x for Int8. The performance gain over TVM is mainly due to the design choice of advanced algorithm, microkernel, and tuned heuristics. oneDNN Graph Compiler inherits these advantages so both have shown significant speedup over TVM on individual matmul kernels. oneDNN Graph compiler and oneDNN primitives performance are at the same level on the matmul operations in general. For the specific MLP tests and shapes, oneDNN Graph Compiler performs 4% worse than primitives on FP32 and 6% better on Int8 when comparing total execution time. We observe performance differences for individual kernels and identified three major reasons. First, oneDNN Graph compiler has the layout propagation optimization which uses blocked layouts for the intermediate tensors produced within the subgraph. oneDNN primitives use the plain layout for these tensors. Second, due to the MLP-1 tests' execution time being relatively short, the total API call overhead takes up to 10% of the execution time. The API call overhead is reduced by about 3 times for oneDNN Graph Compiler since the compiled code only needs to be called once. These two reasons explain oneDNN Graph Compiler's performance gains on small kernels. Lastly, the performance difference is also due to the different choices of parameters made by heuristics and template algorithms.

Fig. 8 shows the performance speedup of oneDNN Graph Compiler on MLP and MHA tests. The default configuration has coarse-grain fusion, and we add an ablation study to understand its benefit. The performance results show that oneDNN Graph Compiler demonstrates an average of 2.4x speedup over TVM on MLP and 6.2x on MHA. Coarse-grain fusion contributes to an average performance gain of 1.2x on both MLP and MHA tests. Most of MLP performance gain can be explained by the kernel performance difference and coarse-grain fusion, MHA has extra performance gain due to oneDNN Graph Compiler's fine-grain fusion optimization. TVM outperforms oneDNN Graph Compiler for the first MLP_1 test, mainly due to that TVM has better-performing kernels for this test.

Coarse-grain fusion accounts for 2.0x for Int8 tests and 3%x for FP32 tests on MLP-1 test, but only 6% for Int8 and 2% for FP32 on MLP-2 test. For MLP-1 Int8 test, coarse-grain fusion is able to merge the outermost parallel loops, lowered from 3 matmul ops, into one parallel loop. The coarse-grain fusion greatly reduces the synchronization overhead and permits the activation data to be in the fastest cache for the next matmul op. As the entire activation and weight tensor fit in the L2 cache, so the coarse-grain fusion gives a much higher speed up for MLP-1 Int8 tests. Other MLP tests also benefit from the coarse-grain fusion but to a lesser extent, since coarse-grain fusion is not able to merge all the loop nests due to the current heuristic limitation.

For the MHA subgraph, the 6.2x performance gain over TVM is from individual kernel performance, fine-grain fusion, and coarse-grain fusion. TVM doesn't fuse the softmax op with the preceding batch matmul op, while oneDNN Graph Compiler decomposes softmax op to multiple basic operations and fuses them to the preceding batch matmul ops with fine-grain fusion Besides, oneDNN Graph Compiler uses a fast implementation of softmax, removing a max reduction while not impacting Bert model accuracy. On top of fine-grain fusion, coarse-grain fusion merges the two nested loops lowered from two batch matmul ops.

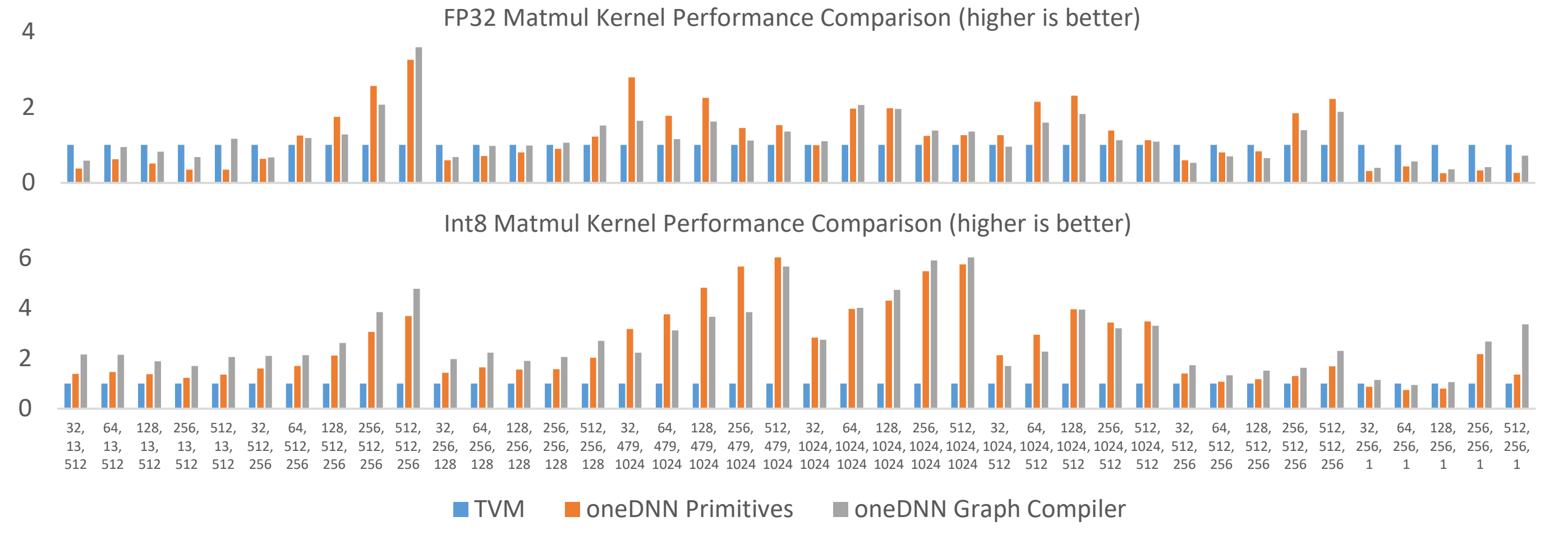

Fig. 7. Matmul kernel execution time comparison between oneDNN primitives, TVM, and oneDNN Graph Compiler

With coarse-grain fusion and fine-grain fusion optimization, oneDNN Graph Compiler outperforms oneDNN post-op fusion for both MLP and MHA tests. For MLP, oneDNN Graph Compiler is on par with oneDNN post-op fusion for FP32 tests' total execution time and demonstrates a 23% speedup for Int8. The coarse-grain fusion helps to recover the kernel performance difference in FP32 and contributes more to Int8 cases since the small kernels account for a higher ratio in Int8 test cases. For MHA, oneDNN Graph Compiler reduces the total execution time by 2.3x on FP32 and 2.4x on Int8.

To demonstrate the end-to-end model performance, we perform inference mode benchmarking for BERT Large and DLRM. We test both Int8 and FP32 data type and tried various batch sizes. We use Intel Extension for Pytorch [29], which offloads MHA and MLP to oneDNN Graph Compiler through oneDNN Graph API. We are not able to show end-to-end model performance for TVM using the pure TVM compilation approach due to the excessive auto-scheduler search time. Since it takes multiple hours to search for optimal kernel implementation for one subgraph, it is impractical for us to conduct such an experiment at the DNN model level.

The performance result in Fig. 9 shows oneDNN Graph Compiler improves BERT_Large throughput by an average of 1.12x and DLRM by 1.15x over oneDNN post-op fusion. Particularly, oneDNN Graph compiler improves Int8 BERT_Large throughput by 1.18x for batch size 128 and Int8 DLRM by 1.21x for batch size 32.

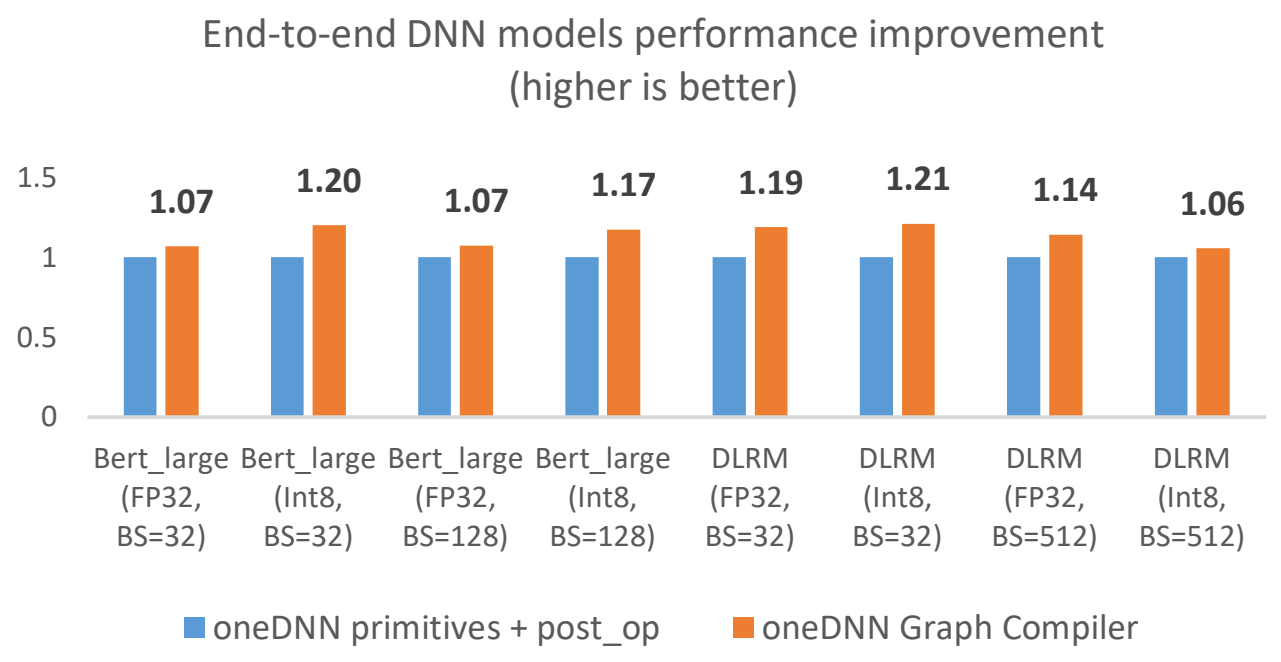

Fig. 9. End-to-end model speedup by oneDNN Graph Compiler

## VIII. Conclusion

We propose a hybrid approach to address the unique challenges of deep learning compilation. It distills key ingredients of expert-tuned primitives for compute-intensive DNN operations like matrix multiplication and uses compiler techniques on the DNN computation graph to fully exploit the performance opportunity at the graph level. The template uses an expert-developed microkernel, algorithm, and heuristic, to ensure compiler-generated code achieves comparable performance to expert-tuned primitives. The compiler uses two-level intermediate representations at the level of both DNN op graph and C program to support domain-specific optimizations needed for deep learning computation, including low-precision, constant weight, tensor memory layout, fine-grain fusion, coarse-grain fusion, and tensor memory buffer reuse. Performance evaluation shows significant performance gain over existing tensor compiler and primitives library for performance critical DNN computation graph and end-to-end models in CPU inference usage.

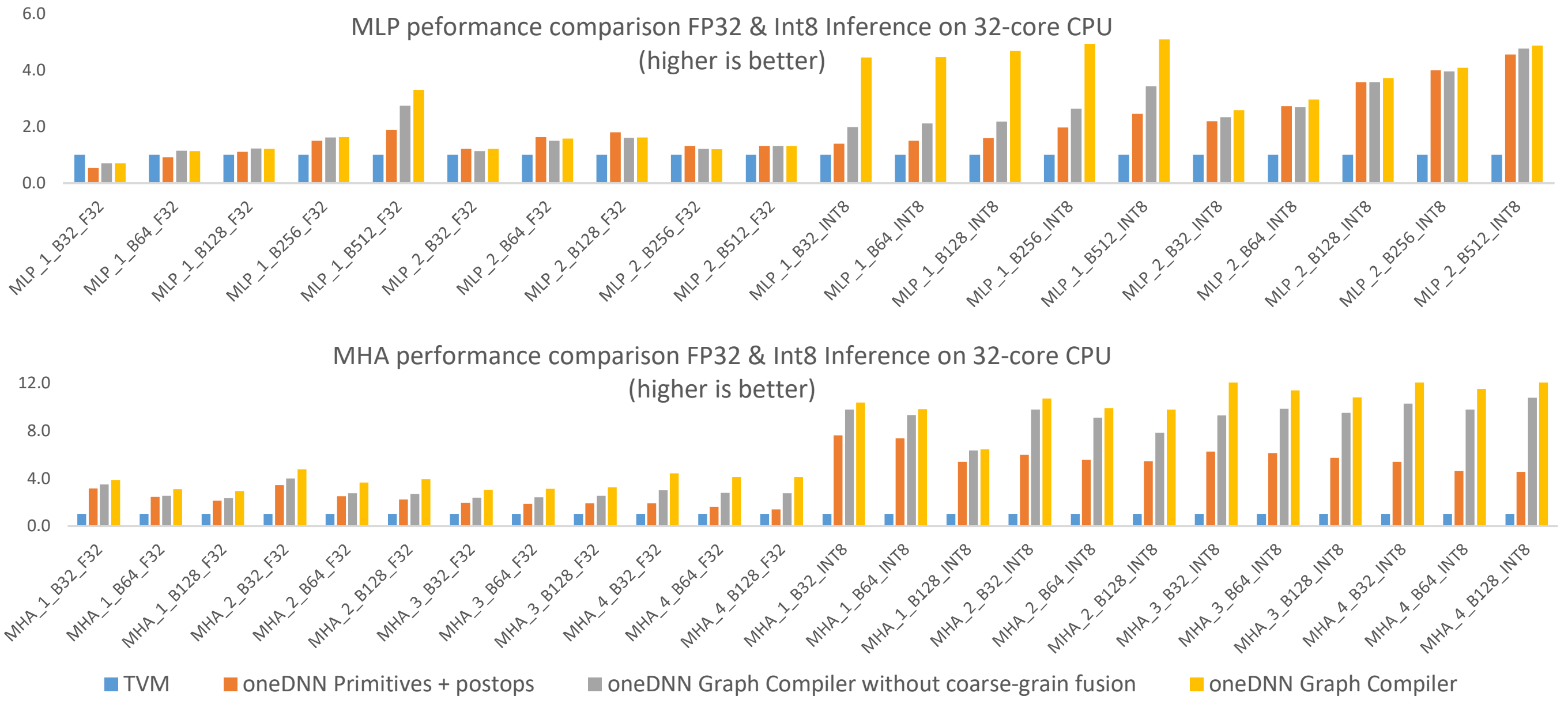

Fig. 8. oneDNN Graph Compiler performance evaluation for MLP and MHA subgraph